\pdfoutput=1

\documentclass[11pt]{article}

\usepackage{acl}

\usepackage{authblk}
\usepackage{times}
\usepackage{latexsym}

\usepackage[T1]{fontenc}

\usepackage[utf8]{inputenc}

\usepackage{microtype}

\usepackage{amsmath}
\usepackage{mathtools}
\usepackage{tabularx}
\usepackage{multirow}
\usepackage{numprint} 
\usepackage{xcolor}
\usepackage{booktabs}
\usepackage{makecell}
\usepackage{longtable}
\usepackage{comment}
\usepackage{float}

\title{Self-Repetition in Abstractive Neural Summarizers}

\author[1]{\textbf{Nikita Salkar}}
\author[2]{\textbf{Thomas Trikalinos}}
\author[1]{\textbf{Byron C. Wallace}}
\author[3]{\textbf{Ani Nenkova}}

\affil[1]{\footnotesize Khoury College of Computer Sciences, Northeastern University, USA}
\affil[2]{\footnotesize Health Services, Policy and Practice, Brown University, USA}
\affil[3]{\footnotesize Adobe Research, USA}
\affil[ ]{\footnotesize \{salkar.n,b.wallace\}@northeastern.edu, thomas\_trikalinos@brown.edu, nenkova@adobe.com}

\begin{document}
\maketitle

\begin{abstract}
We provide a  quantitative  and qualitative analysis of self-repetition in the output of neural summarizers. 
We measure self-repetition as the number of $n$-grams of length four or longer that appear in multiple outputs of the same system. We analyze the behavior of three popular architectures (BART, T5 and Pegasus), fine-tuned on five datasets. 
In a regression analysis, we find that the three architectures have different propensities for repeating content across output summaries for inputs, with BART being particularly prone to self-repetition. 
Fine-tuning on more abstractive data, and on data featuring formulaic language,
is associated with a higher rate of self-repetition. 
In qualitative analysis we find systems produce artefacts such as ads and disclaimers unrelated to the content being summarized, as well as formulaic phrases common in the fine-tuning domain. 
Our approach to corpus level analysis of self-repetition may help practitioners clean up training data for summarizers and ultimately support methods for minimizing the amount of self-repetition. 
\end{abstract}

\section{Introduction}

Sequence-to-sequence neural models for conditional text generation such as BART~\cite{lewis2019bart}, T5~\cite{raffel2020exploring}, and Pegasus~\cite{zhang2020pegasus} 
achieve strong empirical results on abstractive summarization tasks.
The summaries that such systems output often appear to be novel, in that they repeat text verbatim from inputs sparingly or not at all. 
Here, we set out to study the novelty of models with respect to their own outputs, 
by measuring the extent to which the content a model generates is formulaic repetition produced \emph{across inputs}. 

More specifically, we analyze how often long $n$-grams (length $\geq$4)  appear in at least two summaries for different inputs. 
Repetition of some such $n$-grams may be natural, 
for example in news 
covering the same type of event, or in academic papers with accepted formulaic descriptions of research questions and findings. 
To contextualize our measurements, we therefore contrast repetition in summaries written by humans with what we observe in system outputs. 
The former provides a baseline expectation regarding how much repetition is normal in a particular domain. 
In three out of the five domains we study we find that long $n$-gram repetition is considerably higher in automatically produced summaries than in human-written summaries. In the fourth domain, scientific papers, self-repetition even in human summaries is so high that the measure we use may not be sensitive enough to distinguish differences in repetition at this range.


We hypothesized that such undesirable behavior can be easier to quantify when we evaluate systems across domains, tasking a system trained in one domain to generate summaries in another. The intuition was that the repeated $n$-grams will be typical for the fine-tuning domain but rare in the test domain, so problematic repetitions may be easier to detect. 
This setting leads to clear cases of hallucinations reflecting the training data, e.g., 
fine-tuning BART \cite{lewis2019bart} on an academic paper summarization dataset and then applying it to a news summarization task yields hundreds of generated summaries that contain  
the phrase {\em this paper reports the results of an investigation}.
Further, the phrase {\em The past few years have seen a dramatic increase} appears in a dozen news summaries, as do slight variations. 
Table~\ref{table:examples} shows more examples of self-repetition and Section \ref{qualitative} describes the details of our qualitative analysis of $n$-grams identified by manually scanning repeated $n$-grams that clearly do not match the domain of text for which the summaries were generated.

To characterize this repetition behavior quantitatively, we perform a regression analysis in which we include as predictors system architecture,
as well as training and test datasets (Section \ref{section:regression-analysis}). 
We find that BART \cite{lewis2019bart} is especially prone to self-repetition, more so than the other architectures we consider, and that the type of training data used to fine tune the sequence-to-sequence model for summarization 
has a considerable impact on the propensity of models to repeat themselves. 


Our work highlights a dimension of repetition and novelty in summarization that, to our knowledge, has not been explored previously. 
The repetition metrics we introduce may be broadly useful in characterizing the performance of new abstractive summarization systems, as we show that models differ markedly with respect to these measures. 

\begin{table}
\small
\begin{tabular}{@{}ll@{}}
\cmidrule(r){1-2}

\multicolumn{1}{l|}{\parbox[t]{5.5cm}{\textbf{Repeating n-gram}}} & \textbf{Freq}\\ \cmidrule(r){1-2}
\cmidrule(r){1-2}

\multicolumn{1}{l|}{{\parbox[t]{5.5cm}{click here for all the latest transfer news}}} & 73/11490 \\

\cmidrule(r){1-2}

\multicolumn{2}{l}{\begin{tabular}[c]{@{}l@{}}\parbox[t]{7.4cm}{\textcolor{blue}{Example: } Moha El Ouriachi is set to sign for Stoke City, according to his agent. The 19-year-old Barcelona B player is keen to seek first-team action. Stoke have already signed Bojan Krkic and Marc Muniesa from Barcelona. \textcolor{red}{Click here for all the latest transfer news}.} \end{tabular}} \\ 

\cmidrule(r){1-2}

\multicolumn{1}{l|}{\parbox[t]{5.5cm}{this paper reports the results of an investigation}} & 143/11490\\ 

\cmidrule(r){1-2}

\multicolumn{2}{l}{\parbox[t]{7.4cm}{\textcolor{blue}{Example:} schoolgirl killer Zbigniew Huminski was arrested for a range of crimes which are likely to see him jailed for life . \textcolor{red}{this paper reports the results of an investigation} into the circumstances under which he was arrested in the northern port city of Calais} }  \\

\cmidrule(r){1-2}

\multicolumn{1}{l|}{{\parbox[t]{5.5cm}{In our series of letters from African-American journalists, film-maker and columnist Farai Sevenzo considers}}} & 16/11490 \\

\cmidrule(r){1-2}

\multicolumn{2}{l}{\begin{tabular}[c]{@{}l@{}}\parbox[t]{7.4cm}{\textcolor{blue}{Example: } \textcolor{red}{In our series of letters from African-American journalists, film-maker and columnist Farai Sevenzo considers} the lessons learned from the 2013 Boston Marathon bombings.} \end{tabular}} \\ 

\cmidrule(r){1-2}

\multicolumn{1}{l|}{{\parbox[t]{5.5cm}{however, there is insufficient evidence to}}} & 1086/6440 \\

\cmidrule(r){1-2}

\multicolumn{2}{l}{\begin{tabular}[c]{@{}l@{}}\parbox[t]{7.4cm}{\textcolor{blue}{Example: }@xmath3 is an effective solution for the vacuum state of qcd . \textcolor{red}{However, there is insufficient evidence to} support or refute the use of lattice simulations with @xmath3.} \end{tabular}} \\ 

\cmidrule(r){1-2}

\multicolumn{1}{l|}{{\parbox[t]{5.5cm}{but there is a lack of evidence to support}}} & 103/6440 \\

\cmidrule(r){1-2}

\multicolumn{2}{l}{\begin{tabular}[c]{@{}l@{}}\parbox[t]{7.4cm}{\textcolor{blue}{Example: } The Apple Watch is officially going on sale -  \textcolor{red}{but there is a lack of evidence to support} its decision to make it available through online orders.} \end{tabular}} \\ 

\cmidrule(r){1-2}

\end{tabular}
\caption{Examples of self-repetition.}
\label{table:examples}
\end{table}

\section{Related work}
Prior work in abstractive neural summarization has focused on phrases repeated \emph{within a given output}, and proposed various means for mitigating this problem \cite{see2017get, paulus2017deep,fu2021theoretical, nair2021reducing}. 
By contrast, our work quantifies the extent to which systems produce \emph{the same $n$-grams across different inputs}, and the factors that correlate with this behavior. 

Research in text generation has documented that systems often self-repeat and have quantified how much models repeat content from their pre-training data~\cite{mccoy2021much,carlini2022quantifying}. We provide some puzzling examples where we are unable to trace the origin of repeated content\footnote{Recently developed techniques for attributing content in a summary to the language model or the input~\cite{xu-durrett-2021-dissecting} would be more powerful than the manual inspection we carried out and will support future work on self-repetition.}. We also recognize a portion of the repetitions as hallucinations that are influenced by the training data. Oftentimes, the hallucinations are stylistic, similar to the formulaic phrases from academic papers that we mentioned in the introduction. Prior work has shown that neural summarization systems are capable of choosing important content across domains but need in-domain data to faithfully reproduce the style of a given domain~\cite{hua-wang-2017-pilot}. In our work, we find that once systems pick up stylistic templates from one domain, they are likely to reuse them in other domains, where the formulaic phrases look out of place. 

Self-repetition is well-documented in dialog systems research.
Dialog systems often produce generic formulaic responses regardless of the preceding utterance~\cite{li-etal-2016-diversity}: in one of the reported experiments, four generic responses ({\it I don't know, I don't know what you are talking about, I don't think this is a good idea, Oh my god}) constitute 32\% of system generated responses. These phrases were common in the training data, with 0.4\% of training data sentences containing the phrase {\it I don't know}, even though overall the training data was diverse. Our findings for summarization are similar, as we discover in our regression analysis that training on data with higher incidence of formulaic phrases, like academic papers and summaries of medical evidence, results in a summarizer that is overall more likely to repeat content across inputs, at rates markedly higher than done by humans. 

Human summaries are typically considered an appropriate reference while enhancing abstractive text summarization models~\cite{yang2019exploring, yang2020hierarchical}. For our analysis too, we contrast model generated summaries against the human summaries as baseline to determine the threshold over which self--repetition is considered anomalous.

\section{Defining Self-Repetition}
\label{section:self-repetition}

We introduce a \emph{repetition score} to measure how often systems repeat themselves. 
The score is a function of $n$-grams of length four and longer in different summaries, which is indicative of text similarity and potential pliagiarism~\cite{lyon-etal-2001-detecting}.
We consider an $n$-gram to be \textit{repeating} when it appears in two or more summaries in a dataset.
The repetition score can be computed at the \emph{dataset} and \emph{individual summary} level. 

At the dataset level, we count the number of summaries that contain at least one $n$-gram (n$\geq$4) that also appears in another summary. 
We define the repetition score for a dataset as the number of summaries containing repeating \emph{n}-grams divided by the total number of summaries in that dataset. We divide by the total in order to normalize the values allowing for meaningful comparison between datasets of different sizes.

For an individual summary, we define the repetition score as:

\begin{equation}
    R_i = \log(\sum_{k=1}^{m} N_k + 1).\label{eq:sumrep} 
\end{equation}

\noindent Where $i$ indexes summaries, $m$ is the number of repeating $n$-grams in summary $i$, and $N_k$ denotes the count of summaries that contain the $k$th repeating $n$-gram found within summary $i$.
We take the log to this value to produce the final score, to make the repetition score less sensititive to outliers. 


\section{Models and Datasets} 
We consider three models: BART~\cite{lewis2019bart}, T5~\cite{raffel2020exploring}, and Pegasus~\cite{zhang2020pegasus}, 
each fine-tuned on five summarization datasets: CNN/DailyMail~\cite{hermann2015teaching}, BBC XSum ~\cite{narayan2018don}, Scientific Papers (SP; \citealt{cohan2018discourse}), Reddit~\cite{volske-etal-2017-tl} and a corpus of Randomized Controlled Trials (RCTs; ~\citealt{AMIA-summarization-2021}). 
We evaluate each model on the five datasets, yielding 75 (3$\cdot$5$\cdot$5) combinations of architectures, train, and test datasets. 

Table \ref{table:repetition-scores} reports repetition scores for each architecture on the datasets considered.  
To contextualize these, we also report repetition scores for the reference (i.e., human-written) summaries.  
Reddit 
shows the least amount of human repetition; only 27\% of summaries contain at least one $n$-gram of length four or greater 
that also appears in another Reddit summary. 
Scientific Papers are the most formulaic: 99\% of abstracts contain such repetition.  
The RCTs data (also scientific in nature) is similarly repetitive. 
News---from both CNN/Daily Mail and XSum---is somewhere in-between:  
60--70\% of human summaries contain a long repeated $n$-gram.

In model outputs we observe a level of repetition similar to what is seen in the references on the Reddit and Scientific Papers dataset. 
For news corpora (CNN and XSum) and the medical evidence summarization task (RCTs) however, system repetition scores are markedly higher than the scores for the human-written summaries.  
BART seems particularly prone to repetition. 

\begin{table}
\small
\centering
\resizebox{\columnwidth}{!}{%
\begin{tabular}{@{}lrrrr@{}}
 \hline
  \textbf{Dataset} & \textbf{Human} & \textbf{BART} & \textbf{T5} & \textbf{Pegasus}\\ 
  \hline
 CNN/DailyMail & 0.69 & 0.96 & 0.90 &  0.80\\ 
 XSum & 0.60 & 0.85 & 0.70 &  0.81\\  
 Reddit & 0.27 & 0.26 & 0.28 & 0.29\\
 Scientific Papers & 0.99 & 0.99 & 0.99 &  0.99\\ 
 RCT & 0.88 & 1.0 & 0.96 &  1.0\\
 \hline
\end{tabular}%
}
\caption{Repetition scores for human and in-domain system summaries produced with different architectures.}
\label{table:repetition-scores}
\end{table}


We contrast the repetition score of the human summaries in each domain with their level of \emph{abstractiveness}, defined as 
the fraction of $n$-grams of a given size that \emph{do not} appear in the input (and so are ``novel'').
As pointed out in ~\cite{narayan2018don}, reference summaries in XSum are more abstractive than those in the CNN/Daily Mail dataset. 
Table \ref{table:human-summary-novelty} also highlights that  Reddit summaries are particularly extractive, e.g., bi-grams in references almost always appear in the corresponding inputs. 
Aside from Reddit, the number of novel with respect to the input $n$-grams increases with $n$.

\begin{table}
\small
\centering
\resizebox{\columnwidth}{!}{%
\begin{tabular}{@{}lrrrr@{}}
 \hline
  \textbf{Dataset} & \textbf{Unigram} & \textbf{Bigram} & \textbf{Trigram} & \textbf{4-gram}\\ 
  \hline
 CNN/DailyMail & 30.20 & 54.40 & 71.53 & 79.99\\ 
 XSum & 40.40 & 81.47 & 91.47 &  93.64\\  
 Reddit & 9.50 & 2.71 & 2.53 & 2.77\\
 SP & 48.41 & 49.99 & 70.08 & 81.48\\ 
 RCT & 52.56 & 77.87 & 92.02 & 96.08\\
 \hline
\end{tabular}%
}
\caption{Percent abstractiveness of human summaries.}
\label{table:human-summary-novelty}
\end{table}

\section{Qualitative Analysis}
\label{qualitative}

To glean a qualitative view of repetition behavior, we randomly sampled 20 long $n$-grams that appeared in more than 10 summaries.
These $n$-grams often do not appear in the corresponding inputs. 

We show examples in Table \ref{table:examples}. 
The first $n$-gram is generated in 73 out of 11,490 summaries by a Pegasus model fine-tuned on CNN/Daily Mail and applied to test instances from the same domain; there is no domain shift here. This $n$-gram does not occur in the train or the test set.

Repetition is particularly pronounced when the model is trained to summarize data for one domain and then applied to another. 
For example, the second $n$-gram shown (\textit{"this paper reports the results of an investigation"}) was repeated in 143/11,490 summaries generated by a BART model trained on Scientific Papers and then applied to CNN/DailyMail inputs. This $n$-gram also appears in two out of 203,037 training inputs of Scientific Papers with its sub-$n$-grams appearing with even greater frequency.

The next $n$-gram is found in 16 out of 11,490 summaries produced by a BART model trained on XSum and applied to CNN/Daily Mail. This $n$-gram does not appear in the XSum train set; moreover, there is no mention of ``Farai Sevenzo'' in the CNN dataset at all (inputs or outputs).
While these examples contain summaries that are at least related to the input,  Table \ref{table:hallucinations} shows examples of hallucinating summaries generated by Pegasus trained on XSum and then applied to Reddit. The $n$-gram “In our series of letters from African journalists, filmmaker and columnist” occurs in the generated summaries without having any relevance to the input.
These examples indicate that models sometimes produce formulaic content unrelated to inputs, which may not even have been encountered in the training data. 
 

\begin{table}
\small
\centering
\resizebox{0.99\columnwidth}{!}{%
\begin{tabular}{l}
 \hline

 \parbox[t]{7cm}{\vspace{0.05em}\textcolor{red}{Summary}: In our series of letters from African journalists, filmmaker and columnist  Ahmedou Ould-Abdallah reflects on his time at the University of Cape Town.\vspace{0.6em}}\\ 
 
 \parbox[t]{7cm}{\textcolor{blue}{Input} :This is great. I hope you discuss some of these points in the next episode of TLDR: CLG.\vspace{0.4em}}\\

 \hline
 
 \parbox[t]{7cm}{\vspace{0.05em}\textcolor{red}{Summary}: In our series of letters from African journalists, filmmaker and columnist GustavoM looks at the relationship between humans and animals.\vspace{0.6em}}\\ 
 
 \parbox[t]{7cm}{\textcolor{blue}{Input}: Listen to GustavoM, a friendship could never come close to that of a companion.tl;dr: GustavoM is right.\vspace{0.4em}}\\

 \hline
 
 \parbox[t]{7cm}{\vspace{0.05em}\textcolor{red}{Summary}: In our series of letters from African journalists, filmmaker and columnist Ahmedou Ould-Abdallah reflects on his time at the University of Cape Town.\vspace{0.6em}}\\ 
 
 \parbox[t]{7cm}{\textcolor{blue}{Input}: We had to take business writing classes to graduate in finance. It was essentially a class on how to do effective TL;DRs.\vspace{0.4em}}\\

 
 

 \hline
\end{tabular}%
}
\caption{Examples of hallucinations in summaries.}
\label{table:hallucinations}
\end{table}

\section{Regression Analysis}
\label{section:regression-analysis}

We next quantify the association between self-repetition and factors that might influence this, including system architecture and pre-training, and the datasets used for training and testing. 
We would also expect that repetition would be 
proportional to summary length: More words naturally afford more opportunities for repetition, even if by chance. 
And indeed we observe that the repetition scores of human summaries are proportional to their average lengths. We report summary lengths in Appendix Table A1
which can be compared to the repetition scores in Table \ref{table:repetition-scores}. 
Model generated summaries exhibit a similar correlation. 


We also hypothesized that domain shift --- e.g., testing a model trained to summarize scientific texts on news articles --- would increase repetition across summaries (the model may default to stock phrases in such cases). 
We provide qualitative examples of this in Section \ref{qualitative}. 


We fit a regression model to 731,406 summaries generated by 75 combinations of architecture, train and test data, along with the reference summaries for all datasets.
We have multiple one-hot encoded categorical variables, which means we must select reference categories for these (effectively the intercept term). We use human generated summaries as the reference architecture and the CNN/Daily Mail as the reference train and test sets.

\begin{table}[t]
\large
\resizebox{\columnwidth}{!}{%
\npdecimalsign{.}
\nprounddigits{2}
\begin{tabular}{ln{2}{2}n{3}{2}n{3}{2}n{3}{2}}
\hline
                             & \textbf{\hspace{0.2cm}Coef} & \textbf{\hspace{0.25cm}P$> |$t$|$} & \textbf{\hspace{0.3cm}[0.025} & \textbf{\hspace{0.2cm}0.975]}  \\                             
\midrule
\textbf{Intercept}           &       1.9388        &        0.000        &        1.913    &        1.965     \\
\textbf{Length of Summary} &       0.3485  &        0.000        &        0.342    &        0.355     \\
\textbf{BART}                &       1.7933  &      0.000        &        1.770    &        1.817     \\
\textbf{T5}                  &      -0.1103  &      0.000        &       -0.134    &       -0.087     \\
\textbf{Pegasus}             &      -0.0216  &      0.074        &       -0.045    &        0.002     \\
\textbf{Train SP}          &       1.4341  &       0.000        &        1.404    &        1.464     \\
\textbf{Train RCT}         &       2.2816  &       0.000        &        2.251    &        2.312     \\
\textbf{Train Reddit}      &      -0.3654  &       0.000        &       -0.396    &       -0.335     \\
\textbf{Train XSum}        &       0.2353  &       0.000        &        0.204    &        0.266     \\
\textbf{Test SP}            &       0.5538  &       0.000        &        0.518    &        0.590     \\
\textbf{Test RCT}           &      -0.9488  &       0.000        &       -1.062    &       -0.835     \\
\textbf{Test Reddit}        &      -0.5182  &       0.000        &       -0.546    &       -0.490     \\
\textbf{Test XSum}          &      -0.3697  &       0.000        &       -0.400    &       -0.339     \\
\textbf{RCT - SP}          &       2.8972  &        0.000        &        2.845    &        2.949     \\
\textbf{RCT - RCT}         &       2.4053  &        0.000        &        2.254    &        2.556     \\
\textbf{RCT - Reddit}      &       0.4003  &        0.000        &        0.360    &        0.441     \\
\textbf{RCT - XSum}        &      -0.0700  &        0.002        &       -0.114    &       -0.026     \\
\textbf{Reddit - SP}       &       0.5956  &        0.000        &        0.544    &        0.648     \\
\textbf{Reddit - RCT}      &       0.3983  &        0.000        &        0.238    &        0.559     \\
\textbf{Reddit - Reddit}   &      -0.7110  &        0.000        &       -0.752    &       -0.670     \\
\textbf{Reddit - XSum}     &       0.3313  &        0.000        &        0.287    &        0.375     \\
\textbf{SP - SP}           &       0.5053  &        0.000        &        0.454    &        0.556     \\
\textbf{SP - RCT}          &      -0.4508  &        0.000        &       -0.611    &       -0.290     \\
\textbf{SP - Reddit}       &       0.4864  &        0.000        &        0.446    &        0.527     \\
\textbf{SP - XSum}         &       0.1535  &        0.000        &        0.109    &        0.198     \\
\textbf{XSum - SP}         &       0.6605  &        0.000        &        0.608    &        0.713     \\
\textbf{XSum - RCT}        &       0.8076  &        0.000        &        0.647    &        0.968     \\
\textbf{XSum - Reddit}     &       0.4396  &        0.000        &        0.399    &        0.480     \\
\textbf{XSum - XSum}       &       0.1130  &        0.000        &        0.069    &        0.157     \\
\bottomrule
\hline
\end{tabular}%
}
\caption{Regression results; detailed descriptions of predictors are in the Appendix.}
\label{table:regression}
\end{table}

This model treats the repetition observed in a given summary as defined in Equation \ref{eq:sumrep} as a linear function of predictors including:
the length of the generated summary in number of white space delimited tokens (Length of Summary); the model architecture used to generate the summary. Differences in pre-training data will be folded in the behavior due to architecture (BART, T5, Pegasus); the 
training data to which this model was fit; the test data for which a summary is produced; and interaction terms between train and test datasets.
The latter we denote by ``TRAIN - TEST'', e.g., ``XSum - Reddit'' indicates a summary produced by a model fine-tuned on XSum given an input drawn from the Reddit corpus. 
This is a \emph{cross-domain} model. 
By contrast, ``XSum - XSum'' is an \emph{in-domain} example of a summary produced on an XSum test instance by a model fine-tuned using the XSum training data. 
Table \ref{table:regression} enumerates all covariates (more details in the Appendix). 





Table \ref{table:regression} reports results from this analysis. 
We make a few key observations here. 
First, it would seem BART is most prone to repetition of the models considered. 
From the average summary lengths reported in Appendix Table A1, 
we observe that the BART summaries on CNN/DailyMail are almost double the length of human summaries. 
This suggests the possibility that the observed tendency of BART to disproportionately produce repetitions may owe to the fact that it is prone to producing lengthier summaries in general. 
To investigate this, we imposed a restriction on the \textit{max-length} while decoding --- specifically we set this to 50, which falls between the average lengths of T5 and Pegasus of each corresponding model (Appendix Table A2). 
This resulted in BART yielding summaries that are shorter (on average) than those of T5 and Pegasus. 
Table \ref{table:regression with shortened BART} shows the regression results when the analysis performed with these shortened BART summaries. 
This does shrink the coefficient for BART by a small amount, but it remains by far the largest (compared to T5 and Pegasus). This indicates that while the summary length may somewhat influence the overall repetition, BART seems prone to this behavior independent of its tendency to produce lengthier outputs.

In Table \ref{table:regression}, among the source data, RCT has the maximum amount of repetition in comparison to the baseline CNN DailyMail followed by Scientific Papers and XSum, which aligns with the results of Table \ref{table:repetition-scores}. 
Among the test set, Scientific Papers is the only corpus to have an influence on the repetition.
The interaction terms yield higher coefficients when the training data is Scientific Papers or Randomized Controlled Trials in comparison to when the train source is XSum or Reddit. Further, for all the training datasets, the higest values are for when the test data is Scientific Papers or RCT.


To ascertain whether domain shift (in general) is indeed a significant factor associated with repetition, we perform a likelihood ratio test with the interaction terms. 
Specifically we use as our nested model a regression with all  interaction terms omitted, and compare this to the full model with all factors. 
We choose 0.001 as the critical value. 
The likelihood ratio test results in a p-value of $<< 0.001$. 
This implies that the domain interactions do impart information in terms of quantifying the self-repetition, i.e., applying a summarization model to data from a domain that differs from its training source correlates with increased repetition. 


\section{Conclusions}
We evaluated the tendency of neural summarization models to \emph{repeat} themselves \emph{across outputs} on five datasets. To our knowledge this is the first analysis of this phenomenon. Our results indicate that BART has the greatest tendency to self-repeat, and that the training source is a significant factor which may lead to this repetition behavior. 
Adapting a summarization model trained on one domain to another (distinct) domain also correlates significantly with repetition; the model may ``not know what to say'' in such cases, and default to stock phrases from the training data.  
We also found that models sometimes repeat long strings of text that do not contain any references in the corresponding inputs or even the training sets. 
These may originate in pre-training data, but more research into such hallucinations is warranted. 
We hope this analysis will encourage development of methods for mitigating the repetition across summaries and for controlling hallucinations in abstractive neural summarizers.

\section*{Acknowledgements}

We are grateful to Tracy King for her careful reading of and detailed comments on an earlier version of this paper.

This research was supported in part by the National Institutes of Health (NIH) under the National Library of Medicine (NLM) grant 2R01LM012086, and in part by the National Science Foundation (NSF) under grant 2211954.
\bibliography{anthology,custom}

\begin{thebibliography}{22}
\expandafter\ifx\csname natexlab\endcsname\relax\def\natexlab#1{#1}\fi

\bibitem[{Bird et~al.(2009)Bird, Klein, and Loper}]{bird2009natural}
Steven Bird, Ewan Klein, and Edward Loper. 2009.
\newblock \emph{Natural language processing with Python: analyzing text with
  the natural language toolkit}.
\newblock " O'Reilly Media, Inc.".

\bibitem[{Carlini et~al.(2022)Carlini, Ippolito, Jagielski, Lee, Tramer, and
  Zhang}]{carlini2022quantifying}
Nicholas Carlini, Daphne Ippolito, Matthew Jagielski, Katherine Lee, Florian
  Tramer, and Chiyuan Zhang. 2022.
\newblock Quantifying memorization across neural language models.
\newblock In \emph{CoRR, vol. abs/2202.07646}.

\bibitem[{Cohan et~al.(2018)Cohan, Dernoncourt, Kim, Bui, Kim, Chang, and
  Goharian}]{cohan2018discourse}
Arman Cohan, Franck Dernoncourt, Doo~Soon Kim, Trung Bui, Seokhwan Kim, Walter
  Chang, and Nazli Goharian. 2018.
\newblock A discourse-aware attention model for abstractive summarization of
  long documents.
\newblock In \emph{Proceedings of the 2018 Conference of the North {A}merican
  Chapter of the Association for Computational Linguistics: Human Language
  Technologies, Volume 2 (Short Papers)}.

\bibitem[{Fu et~al.(2021)Fu, Lam, So, and Shi}]{fu2021theoretical}
Zihao Fu, Wai Lam, Anthony Man-Cho So, and Bei Shi. 2021.
\newblock A theoretical analysis of the repetition problem in text generation.
\newblock In \emph{Proceedings of the AAAI Conference on Artificial
  Intelligence}, volume~35.

\bibitem[{Hermann et~al.(2015)Hermann, Kocisky, Grefenstette, Espeholt, Kay,
  Suleyman, and Blunsom}]{hermann2015teaching}
Karl~Moritz Hermann, Tomas Kocisky, Edward Grefenstette, Lasse Espeholt, Will
  Kay, Mustafa Suleyman, and Phil Blunsom. 2015.
\newblock Teaching machines to read and comprehend.
\newblock \emph{Advances in neural information processing systems}, 28.

\bibitem[{Hua and Wang(2017)}]{hua-wang-2017-pilot}
Xinyu Hua and Lu~Wang. 2017.
\newblock \href {https://doi.org/10.18653/v1/W17-4513} {A pilot study of domain
  adaptation effect for neural abstractive summarization}.
\newblock In \emph{Proceedings of the Workshop on New Frontiers in
  Summarization}, pages 100--106, Copenhagen, Denmark. Association for
  Computational Linguistics.

\bibitem[{Lewis et~al.(2019)Lewis, Liu, Goyal, Ghazvininejad, Mohamed, Levy,
  Stoyanov, and Zettlemoyer}]{lewis2019bart}
Mike Lewis, Yinhan Liu, Naman Goyal, Marjan Ghazvininejad, Abdelrahman Mohamed,
  Omer Levy, Ves Stoyanov, and Luke Zettlemoyer. 2019.
\newblock Bart: Denoising sequence-to-sequence pre-training for natural
  language generation, translation, and comprehension.
\newblock In \emph{Association for Computational Linguistics (ACL)}.

\bibitem[{Li et~al.(2016)Li, Galley, Brockett, Gao, and
  Dolan}]{li-etal-2016-diversity}
Jiwei Li, Michel Galley, Chris Brockett, Jianfeng Gao, and Bill Dolan. 2016.
\newblock \href {https://doi.org/10.18653/v1/N16-1014} {A diversity-promoting
  objective function for neural conversation models}.
\newblock In \emph{Proceedings of the 2016 Conference of the North {A}merican
  Chapter of the Association for Computational Linguistics: Human Language
  Technologies}, pages 110--119, San Diego, California. Association for
  Computational Linguistics.

\bibitem[{Lyon et~al.(2001)Lyon, Malcolm, and
  Dickerson}]{lyon-etal-2001-detecting}
Caroline Lyon, James Malcolm, and Bob Dickerson. 2001.
\newblock \href {https://aclanthology.org/W01-0515} {Detecting short passages
  of similar text in large document collections}.
\newblock In \emph{Proceedings of the 2001 Conference on Empirical Methods in
  Natural Language Processing}.

\bibitem[{McCoy et~al.(2021)McCoy, Smolensky, Linzen, Gao, and
  Celikyilmaz}]{mccoy2021much}
R~Thomas McCoy, Paul Smolensky, Tal Linzen, Jianfeng Gao, and Asli Celikyilmaz.
  2021.
\newblock How much do language models copy from their training data? evaluating
  linguistic novelty in text generation using raven.
\newblock In \emph{CoRR, abs/2111.09509}.

\bibitem[{Nair and Singh(2021)}]{nair2021reducing}
Pranav Nair and Anil~Kumar Singh. 2021.
\newblock On reducing repetition in abstractive summarization.
\newblock In \emph{Proceedings of the Student Research Workshop Associated with
  RANLP 2021}, pages 126--134.

\bibitem[{Narayan et~al.(2018)Narayan, Cohen, and Lapata}]{narayan2018don}
Shashi Narayan, Shay~B Cohen, and Mirella Lapata. 2018.
\newblock Don't give me the details, just the summary! topic-aware
  convolutional neural networks for extreme summarization.
\newblock In \emph{Proceedings of the 2018 Conference on Empirical Methods in
  Natural Language Processing}.

\bibitem[{Paulus et~al.(2018)Paulus, Xiong, and Socher}]{paulus2017deep}
Romain Paulus, Caiming Xiong, and Richard Socher. 2018.
\newblock A deep reinforced model for abstractive summarization.
\newblock In \emph{6th International Conference on Learning
  Representations,Vancouver,BC,Canada}.

\bibitem[{Raffel et~al.(2020)Raffel, Shazeer, Roberts, Lee, Narang, Matena,
  Zhou, Li, Liu et~al.}]{raffel2020exploring}
Colin Raffel, Noam Shazeer, Adam Roberts, Katherine Lee, Sharan Narang, Michael
  Matena, Yanqi Zhou, Wei Li, Peter~J Liu, et~al. 2020.
\newblock Exploring the limits of transfer learning with a unified text-to-text
  transformer.
\newblock \emph{J. Mach. Learn. Res.}, 21(140):1--67.

\bibitem[{Seabold and Perktold(2010)}]{seabold2010statsmodels}
Skipper Seabold and Josef Perktold. 2010.
\newblock statsmodels: Econometric and statistical modeling with python.
\newblock In \emph{9th Python in Science Conference}.

\bibitem[{See et~al.(2017)See, Liu, and Manning}]{see2017get}
Abigail See, Peter~J Liu, and Christopher~D Manning. 2017.
\newblock Get to the point: Summarization with pointer-generator networks.
\newblock In \emph{Proceedings of the 55th Annual Meeting of the Association
  for Computational Linguistics}.

\bibitem[{V{\"o}lske et~al.(2017)V{\"o}lske, Potthast, Syed, and
  Stein}]{volske-etal-2017-tl}
Michael V{\"o}lske, Martin Potthast, Shahbaz Syed, and Benno Stein. 2017.
\newblock \href {https://doi.org/10.18653/v1/W17-4508} {{TL};{DR}: Mining
  {R}eddit to learn automatic summarization}.
\newblock In \emph{Proceedings of the Workshop on New Frontiers in
  Summarization}, pages 59--63, Copenhagen, Denmark. Association for
  Computational Linguistics.

\bibitem[{Wallace et~al.(2021)Wallace, Saha, Soboczenski, and
  Marshall}]{AMIA-summarization-2021}
Byron~C. Wallace, Sayantan Saha, Frank Soboczenski, and Iain~J. Marshall. 2021.
\newblock {Generating (Factual?) Narrative Summaries of RCTs: Experiments with
  Neural Multi-Document Summarization}.
\newblock In \emph{{Proceedings of AMIA Informatics Summit}}.

\bibitem[{Xu and Durrett(2021)}]{xu-durrett-2021-dissecting}
Jiacheng Xu and Greg Durrett. 2021.
\newblock \href {https://doi.org/10.18653/v1/2021.acl-long.539} {Dissecting
  generation modes for abstractive summarization models via ablation and
  attribution}.
\newblock In \emph{Proceedings of the 59th Annual Meeting of the Association
  for Computational Linguistics and the 11th International Joint Conference on
  Natural Language Processing (Volume 1: Long Papers)}, pages 6925--6940,
  Online. Association for Computational Linguistics.

\bibitem[{Yang et~al.(2020)Yang, Li, Shen, Wu, Zhao, and
  Chen}]{yang2020hierarchical}
Min Yang, Chengming Li, Ying Shen, Qingyao Wu, Zhou Zhao, and Xiaojun Chen.
  2020.
\newblock Hierarchical human-like deep neural networks for abstractive text
  summarization.
\newblock \emph{IEEE Transactions on Neural Networks and Learning Systems},
  32(6):2744--2757.

\bibitem[{Yang et~al.(2019)Yang, Qu, Tu, Shen, Zhao, and
  Chen}]{yang2019exploring}
Min Yang, Qiang Qu, Wenting Tu, Ying Shen, Zhou Zhao, and Xiaojun Chen. 2019.
\newblock Exploring human-like reading strategy for abstractive text
  summarization.
\newblock In \emph{Proceedings of the AAAI Conference on Artificial
  Intelligence}, volume~33.

\bibitem[{Zhang et~al.(2020)Zhang, Zhao, Saleh, and Liu}]{zhang2020pegasus}
Jingqing Zhang, Yao Zhao, Mohammad Saleh, and Peter Liu. 2020.
\newblock Pegasus: Pre-training with extracted gap-sentences for abstractive
  summarization.
\newblock In \emph{International Conference on Machine Learning}, pages
  11328--11339. PMLR.

\end{thebibliography}
\appendix
\section{Appendix}
\label{sec:appendix}

\setcounter{table}{0}
\renewcommand{\thetable}{A\arabic{table}}


\begin{table*}[t]
\centering
\small
\def\mystrut{\rule{0pt}{2\normalbaselineskip}}
\begin{tabularx}{\textwidth}{m{2.5cm}m{3.5cm}m{2cm}m{1.7cm}m{1.1cm}m{1.1cm}m{1cm}}
\toprule
\thead[l]{Dataset} & 
\thead[l]{Train / Val / Test} & 
\thead[l]{Input\\Document} & 
\thead[l]{Human\\Summary} & 
\thead[l]{BART} & 
\thead[l]{T5} & 
\thead[l]{\vspace{-13pt}Pegasus}
\\
\midrule
CNN/Daily Mail & 287113 / 11338 / 11490 & 683.51 & 52.12 & 103.37 & 58.01 & 53.16 \\

XSum & 204045 / 11332 / 11334 & 360.58 & 21.09 & 19.34 & 20.13 & 17.86  \\

Reddit & 67198 / 16800 / 16000 & 222.66 & 21.06 & 19.54 & 21.69 & 22.98 \\ 

Scientific Papers & 203037 / 6436 / 6440 & 5702.14 & 163.13 & 96.52 & 81.09 & 97.37 \\

RCT & 3721 / 464 / 466 & 2689.83 & 68.15 & 22.68 & 58.75 & 39.64\\
\bottomrule
\end{tabularx}
\caption{Average lengths of test inputs, the corresponding human summaries, and model-generated summaries.} 
\label{table:average-length}
\end{table*}

\subsection{Regression Model Details}
\label{sec:regression-details}

The dataset for the regression model comprises 731406 summaries, generated by the 75 (3$\cdot$5$\cdot$5) combinations of architectures, train and test datasets. 
The predictors corresponding to each summary $i$ and the observed repetition score $R_i$ constitutes an ($x_i,y_i$) pair. 
More specifically, ``$x_i$'' is composed of the features of the summary we use in our analysis, which we describe individually below.
Note that some of our predictors (those related to architectures and datasets) are categorical, and so need to be ``one-hot'' encoded. 
In such cases, one option must serve as a reference category with respect to which the remaining coefficients can be interpreted.

Regarding these categorical variables: We analyze four architectures for producing summaries --- including ``Human'' in addition to BART, T5 and Pegasus. 
``Human'' serves as our reference architecture, so we do not have an explicit coefficient for this. 
Similarly, we include five datasets in our analysis; for any summary one dataset will have served as the training source and another as the source of test inputs. 
We use CNN/Daily Mail as the reference category for both of these categorical predictors.


Because we are interested in the effects of applying models trained on one summarization domain to another, we also include ``interaction terms'' that encode pairs of train/test datasets via indicators. 
As such, we one-hot encode all pairwise interactions between our four datasets. 

We estimate coefficients to these predictors given the observed summary data in an Ordinary Least Squares (OLS) linear regression model, as implemented the {\tt statsmodels} (v0.12.2) Python module ~\cite{seabold2010statsmodels}. 

\begin{table}[t]
\large
\resizebox{\columnwidth}{!}{%
\npdecimalsign{.}
\nprounddigits{2}
\begin{tabular}{ln{2}{2}n{3}{2}n{3}{2}n{3}{2}}
\hline
                             & \textbf{\hspace{0.2cm}Coef} & \textbf{\hspace{0.25cm}P$> |$t$|$} & \textbf{\hspace{0.3cm}[0.025} & \textbf{\hspace{0.2cm}0.975]}  \\                             
\midrule
\textbf{Intercept}           &       1.6434  &        0.013     &   122.873  &         0.000         \\
\textbf{Length of Summary} &       0.4295  &        0.003     &   129.302  &         0.000          \\
\textbf{BART}                &       1.6007  &        0.012     &   129.281  &         0.000       \\
\textbf{T5}                  &      -0.1289  &        0.012     &   -10.522  &         0.000          \\
\textbf{Pegasus}             &      -0.0403  &        0.012     &    -3.294  &         0.001         \\
\textbf{Train SP}          &       1.6855  &        0.016     &   105.388  &         0.000        \\
\textbf{Train RCT}         &       2.6032  &        0.016     &   165.887  &         0.000        \\
\textbf{Train Reddit}      &      -0.0855  &        0.016     &    -5.400  &         0.000         \\
\textbf{Train XSum}        &       0.6538  &        0.016     &    41.275  &         0.000        \\
\textbf{Test SP}            &       0.5190  &        0.019     &    27.675  &         0.000      \\
\textbf{Test RCT}           &      -0.7732  &        0.059     &   -13.165  &         0.000      \\
\textbf{Test Reddit}        &      -0.5059  &        0.014     &   -35.133  &         0.000      \\
\textbf{Test XSum}          &      -0.2807  &        0.016     &   -17.840  &         0.000        \\
\textbf{RCT - SP}          &       2.9745  &        0.027     &   110.002  &         0.000      \\
\textbf{RCT - RCT}         &       2.2366  &        0.078     &    28.603  &         0.000        \\
\textbf{RCT - Reddit}      &       0.4091  &        0.021     &    19.497  &         0.000       \\
\textbf{RCT - XSum}        &      -0.1315  &        0.023     &    -5.763  &         0.000       \\
\textbf{Reddit - SP}       &       0.6322  &        0.027     &    23.420  &         0.000        \\
\textbf{Reddit - RCT}      &       0.2534  &        0.083     &     3.045  &         0.002      \\
\textbf{Reddit - Reddit}   &      -0.6041  &        0.022     &   -28.068  &         0.000       \\
\textbf{Reddit - XSum}     &       0.2555  &        0.023     &    11.206  &         0.000        \\
\textbf{SP - SP}           &       0.4482  &        0.026     &    16.968  &         0.000       \\
\textbf{SP - RCT}          &      -1.0333  &        0.083     &   -12.414  &         0.000      \\
\textbf{SP - Reddit}       &       0.5118  &        0.021     &    24.367  &         0.000       \\
\textbf{SP - XSum}         &       0.1051  &        0.023     &     4.605  &         0.000        \\
\textbf{XSum - SP}         &       0.6911  &        0.027     &    25.603  &         0.000        \\
\textbf{XSum - RCT}        &       0.6332  &        0.083     &     7.610  &         0.000        \\
\textbf{XSum - Reddit}     &       0.4326  &        0.021     &    20.626  &         0.000        \\
\textbf{XSum - XSum}       &       0.0092  &        0.023     &     0.407  &         0.684        \\
\bottomrule
\hline
\end{tabular}%
}
\caption{Regression results after restricting length of BART summaries.}
\label{table:regression with shortened BART}
\end{table}
\paragraph{Details about regression predictors} We discuss the individual terms in our regression (coefficients for which are reported in Table \ref{table:regression}) in greater detail below. 
\begin{itemize}
    \item \textbf{Length of Summary} This is the number of words in a summary extracted by the NLTK word tokenizer ~\cite{bird2009natural} . Because lengths are quite variable, we standardize the length using the Z-score normalization. The value of $0.35$ in the analysis suggests a positive correlation between the length of a summary and the amount of repetition which also corroborates our observations from Table \ref{table:average-length} and Table \ref{table:repetition-scores}.
    
    \item \textbf{Human} This denotes the special neural ``architecture'' responsible for generating the reference summaries: Humans. Recall that ``humans'' serve as our reference architecture category for one-hot encoding, so are folded into the intercept term.
    
    \item \textbf{BART} This denotes the summaries generated by the BART architecture \cite{lewis2019bart}. The somewhat large positive coefficient ($1.79$) indicates BART is particularly prone to generating repetitions across its outputs. 
    
    \item \textbf{T5} This denotes the summaries generated by the T5 architecture ~\cite{raffel2020exploring}. Overall, our regression results suggest that in aggregate T5 is about comparable to humans in terms of its tendency to repeat itself in general, although it is also subject to this in domain adaptation settings (as are all models considered).
   
    \item \textbf{Pegasus} This denotes the summaries generated by the Pegasus architecture~\cite{zhang2020pegasus}. The interpretation of the corresponding coefficient here is similar to for T5. 
    
    \item \textbf{Train CNN/Daily Mail} This indicates summaries produced by models \emph{trained} on the CNN/Daily Mail dataset ~\cite{hermann2015teaching}. CNN/Daily Mail serves as our reference for this categorical feature, and so we do not have an explicit coefficient for it. 
    
    \item \textbf{Train SP} Indicates a summary produced by a model \emph{trained} on the Scientific Papers dataset ~\cite{cohan2018discourse}. The positive coefficient ($1.43$) suggests that in aggregate models trained on Scientific Papers are more prone to repeat than those trained on CNN/Daily Mail dataset.
    
    \item \textbf{Train RCT} Indicates a summary produced by a model \emph{trained} on the Randomized Controlled Trials (RCTs) dataset~\cite{AMIA-summarization-2021}. The positive coefficient ($2.28$) suggests that training on this dataset results in comparatively large amount of repetition. 
    
    \item \textbf{Train Reddit} Indicates a summary produced by a model \emph{trained} on the Reddit dataset~\cite{volske-etal-2017-tl}. The small negative coefficient value of -$0.37$ indicates that models trained on Reddit are somewhat less prone to repetition, on average.
    
    \item \textbf{Train XSum} Indicates a summary produced by a model \emph{trained} on the XSum dataset~\cite{narayan2018don}. 
    The small positive coefficient estimate of  $0.24$ implies that models trained on XSum may repeat slightly more than those trained on CNN/Daily Mail, on average.
    
    \item \textbf{Test CNN/Daily Mail} Indicates that the corresponding summary was generated for an instance drawn from the CNN/Daily Mail test set. We again treat this as the reference category. 
    
    \item \textbf{Test SP} Indicates that the corresponding summary was generated for an instance drawn from the Test SP test set. The small positive value of $0.55$ suggests that evaluating models on  Scientific Paper instances correlates with a greater amount of repetition. 
    
    
    \item \textbf{Test RCT} 
    Indicates that the corresponding summary was generated for an instance drawn from the Test RCT test set.
    The negative value of -$0.95$ indicates that when tested on RCT instances, models are slightly less prone to repetition.
    
    \item \textbf{Test Reddit}  Indicates that the corresponding summary was generated for an instance drawn from the Reddit test set.
    The small negative value of -$0.52$ implies that when models are evaluated on Reddit instances they may tend to repeat themselves across summaries comparatively less frequently.
    
    \item \textbf{Test XSum}
    Indicates that the corresponding summary was generated for an instance drawn from the XSum test set.
    The negative coefficient of -$0.37$ implies a slightly lower tendency for repetition when models are tested on instances from the XSum test set. 
    
    
    \item \textbf{RCT -- SP} This denotes a summary produced by a model trained on the RCTs train set and evaluated on Scientific Papers test set; a cross-domain scenario. The estimate coefficient of $2.90$ indicates that this combination of interaction yields a comparatively high amount of repetition. 
    
    \item \textbf{RCT -- RCT} This denotes a summary generated by a model trained and tested on the Randomized Controlled Trials. This is a in-domain scenario. A coefficient of $2.41$ indicates that this combination of interaction also yields a much higher amount of repetition than the baseline train - test combination.
    
    \item \textbf{RCT -- Reddit} This denotes a summary produced by a model trained on Randomized Controlled Trials and evaluated on Reddit. This is again a cross-domain scenario. A coefficient of $0.40$ means that this combination has a negligibly higher self-repetion than the baseline.

\begin{table}[t]
\small
\centering
\resizebox{\columnwidth}{!}{%
\begin{tabular}{@{}llrrr@{}}
 \hline
  \vspace{2pt}\textbf{Train} & \textbf{Test} & \textbf{Bart} & \textbf{T5} & \textbf{Pegasus}\\ 
  \hline
 CNN/Daily Mail & CNN/Daily Mail & 103.37 & 58.00 & 53.16 \\
 & XSum & 65.2 & 45.59 & 44.26 \\
 & SP & 92.33 & 63.43 & 45.05\\
 & Reddit & 91.48 & 44.92 & 52.35\\
 & RCT & 77.63 & 39.78 & 43.99\\\\
 XSum & CNN/Daily Mail & 21.69 & 23.06 & 19.11\\
 & XSum & 19.34 & 20.13 & 17.86\\
 & SP & 22.62 & 25.2 & 20.24\\
 & Reddit & 20.12 & 19.83 & 17.61\\
 & RCT & 22.72 & 20.13 & 19.32\\\\
 SP & CNN/Daily Mail & 69.51 & 83.39 & 95.33\\
 & XSum & 58.42 & 78.28 & 66.72\\
 & SP & 96.51 & 81.09 & 97.37\\
 & Reddit & 56.53 & 71.79 & 78.41\\
 & RCT & 83.46 & 46.69 & 66.12\\\\
 Reddit & CNN/Daily Mail & 54.06 & 84.04 & 78.96\\
 & XSum & 53.89 & 78.92 & 69.51\\
 & SP & 62.07 & 69.14 & 83.60\\
 & Reddit & 19.53 & 21.69 & 22.98\\
 & RCT & 44.15 & 46.41 & 92.51\\\\
 RCT & CNN/Daily Mail & 35.16 & 61.95 & 73.53\\
 & XSum & 28.92 & 62.61 & 48.73\\
 & SP & 28.71 & 45.38 & 49.48\\
 & Reddit & 24.60 & 60.40 & 62.59\\
 & RCT & 22.68 & 58.75 & 39.64\\
 \hline
\end{tabular}%
}
\caption{The Average Lengths of Systems before restricting the max-length during BART decoding.}
\label{table:Average Lengths of Systems before restricting BART length.}
\end{table}

\begin{table}[ht]
\small
\centering
\resizebox{\columnwidth}{!}{%
\begin{tabular}{@{}llrrr@{}}
 \hline
  \vspace{2pt}\textbf{Train} & \textbf{Test} & \textbf{Bart} & \textbf{T5} & \textbf{Pegasus}\\ 
  \hline

 CNN/Daily Mail & CNN/Daily Mail & 36.67 & 58.00 & 53.16 \\
 & XSum & 36.61 & 45.59 & 44.26 \\
 & SP & 38.36 & 63.43 & 45.05\\
 & Reddit & 38.52 & 44.92 & 52.35\\
 & RCT & 32.17 & 39.78 & 43.99\\\\
 XSum & CNN/Daily Mail & 21.69 & 23.06 & 19.11\\
 & XSum & 19.34 & 20.13 & 17.86\\
 & SP & 22.62 & 25.2 & 20.24\\
 & Reddit & 20.12 & 19.83 & 17.61\\
 & RCT & 22.72 & 20.13 & 19.32\\\\
 SP & CNN/Daily Mail & 69.51 & 83.39 & 95.33\\
 & XSum & 58.42 & 78.28 & 66.72\\
 & SP & 96.51 & 81.09 & 97.37\\
 & Reddit & 56.53 & 71.79 & 78.41\\
 & RCT & 35.33 & 46.69 & 66.12\\\\
 Reddit & CNN/Daily Mail & 54.06 & 84.04 & 78.96\\
 & XSum & 53.89 & 78.92 & 69.51\\
 & SP & 62.07 & 69.14 & 83.60\\
 & Reddit & 19.53 & 21.69 & 22.98\\
 & RCT & 44.15 & 46.41 & 92.51\\\\
 RCT & CNN/Daily Mail & 35.16 & 61.95 & 73.53\\
 & XSum & 28.92 & 62.61 & 48.73\\
 & SP & 28.71 & 45.38 & 49.48\\
 & Reddit & 24.60 & 60.40 & 62.59\\
 & RCT & 22.68 & 58.75 & 39.64\\
 \hline
\end{tabular}%
}
\caption{The Average Lengths of Systems after restricting the max-length during BART decoding.}
\label{table:Average Lengths of Systems after restricting BART length.}
\end{table}
    
    Similarly, for the rest.
    
    \item \textbf{RCT -- XSum} Denotes a summary generated by a model trained on Randomized Controlled Trials and tested on XSum.
    
    \item \textbf{SP -- SP} Denotes a summary generated by an in-domain model trained and tested on Scientific Papers.
    
    \item \textbf{SP -- RCT} Denotes a summary produced by a trained on Scientific Papers and tested on Randomized Controlled Trials.
    
    \item \textbf{SP -- Reddit} Denotes a summary produced by a model trained on Scientific Papers and tested on Reddit.
    
    \item \textbf{SP -- XSum} Denotes a summary produced by a model trained on Scientific Papers and tested on XSum. 
    
    \item \textbf{Reddit -- SP} Denotes a summary generated by a model trained on Reddit and tested on Scientific Papers.
    
    \item \textbf{Reddit -- RCT} Denotes a summary produced by a model trained on Reddit and tested on Scientific Papers.
    
    \item \textbf{Reddit -- Reddit} Denotes a summary produced by an in-domain model trained and tested on Reddit.
    
    \item \textbf{Reddit -- XSum} Denotes a summary produced by a model trained on Reddit and tested on XSum.
    
    \item \textbf{XSum -- SP} Denotes a summary generated by a model trained on XSum and tested on Scientific Papers.
    
    \item \textbf{XSum -- RCT} Denotes a summary produced by a model trained on XSum and tested on RCT.
    
    \item \textbf{XSum -- Reddit} Denotes a summary produced by a model trained on XSum and tested on Reddit.
    
    \item \textbf{XSum -- XSum} Denotes a summary produced by an in-domain model trained and tested on XSum.
\end{itemize}

Table \ref{table:Average Lengths of Systems before restricting BART length.} reports the average lengths of summaries generated by each system. We can see that when the training data is CNN/Daily Mail, BART has the highest average lengths. Further BART trained on Scientific Papers and applied to RCTs also have lengths higher than corresponding models.

We restrict the max-lengths of these systems to 50 which lies between the corresponding T5 and Pegasus models' average lengths. Table \ref{table:Average Lengths of Systems after restricting BART length.} depicts the average lengths after imposing the restrictions.
From \ref{table:regression with shortened BART} we can see that shortening the lengths of BART summaries does not mitigate its tendency to repeat the most of all the models.

\end{document}